\documentclass{article}
\usepackage{iclr_arxiv,times}

\usepackage{amsmath,amsfonts,bm}

\def\eqref#1{equation~\ref{#1}}
\def\1{\bm{1}}

\DeclareMathAlphabet{\mathsfit}{\encodingdefault}{\sfdefault}{m}{sl}
\SetMathAlphabet{\mathsfit}{bold}{\encodingdefault}{\sfdefault}{bx}{n}

\usepackage{hyperref}
\usepackage{url}

\title{CUDA-Harness: Harnessing Agentic CUDA Kernel Generation and Optimization from Natural Language}

\author{
Qi Fan \& An Zou \& Yehan Ma\thanks{Corresponding Author} \\
Shanghai Jiao Tong University \\
\texttt{\{fanqi666, an.zou, yehanma\}@sjtu.edu.cn} \\
}

\usepackage{amsmath}
\usepackage{amssymb}
\usepackage{graphicx}
\usepackage{booktabs}
\usepackage{multirow}
\usepackage{subcaption}
\usepackage{tikz}
\usepackage{wrapfig}
\usepackage{xargs}
\usepackage{listings}
\usepackage[most]{tcolorbox}
\tcbuselibrary{listings,breakable}

\definecolor{LightBg}{RGB}{250,250,250}
\definecolor{LightFrame}{RGB}{191,191,191}
\definecolor{LightText}{RGB}{30,30,30}
\definecolor{LightKeyword}{RGB}{0,0,255}
\definecolor{LightComment}{RGB}{0,128,0}
\definecolor{LightString}{RGB}{163,21,21}
\definecolor{LightNumber}{RGB}{9,134,88}
\definecolor{LightType}{RGB}{43,145,175}
\definecolor{LightPreproc}{RGB}{128,0,128}
\definecolor{LightLineNo}{RGB}{150,150,150}

\lstdefinelanguage{CUDA}{
  language=C++,
  morekeywords={
    __global__,__device__,__host__,__shared__,__constant__,
    __managed__,__restrict__,__syncthreads,__syncthreads,
    __threadfence,__threadfence_block,__threadfence_system,
    __launch_bounds__,blockIdx,blockDim,threadIdx,gridDim,
    warpSize,cudaMalloc,cudaFree,cudaMemcpy,cudaMemset,
    cudaDeviceSynchronize,atomicAdd,atomicSub,atomicExch,
    atomicMin,atomicMax,atomicInc,atomicDec,atomicCAS,
    atomicAnd,atomicOr,atomicXor
  },
  sensitive=true,
  morecomment=[l]{//},
  morecomment=[s]{/*}{*/},
  morestring=[b]",
  morestring=[b]'
}

\lstdefinestyle{cudaStyleLight}{
  language=CUDA,
  backgroundcolor=\color{LightBg},
  basicstyle=\ttfamily\scriptsize\color{LightText},
  keywordstyle=\bfseries\color{LightKeyword},
  commentstyle=\itshape\color{LightComment},
  stringstyle=\color{LightString},
  identifierstyle=\color{LightText},
  numberstyle=\tiny\color{LightLineNo},
  numbers=left,
  stepnumber=1,
  numbersep=8pt,
  showstringspaces=false,
  showspaces=false,
  showtabs=false,
  tabsize=2,
  breaklines=false,
  breakatwhitespace=false,
  prebreak={},
  postbreak={},
  columns=fullflexible,
  keepspaces=true,
  upquote=true,
  literate={"}{{\char`\"}}1 {'}{{\char`\'}}1
}

\newtcblisting{cudacode}{
  listing only,
  colback=LightBg,
  colframe=LightFrame,
  boxrule=0.5pt,
  arc=1mm,
  left=1mm,
  right=1mm,
  top=0.5mm,
  bottom=0.5mm,
  listing options={
    style=cudaStyleLight
  }
}

\newtcblisting{cudacodearg}[1][]{
  listing only,
  colback=LightBg,
  colframe=LightFrame,
  boxrule=0.5pt,
  arc=1mm,
  left=1mm,
  right=1mm,
  top=0.5mm,
  bottom=0.5mm,
  listing options={
    style=cudaStyleLight,
    #1
  },
}

\newcommand*\circled[1]{\tikz[baseline=(char.base)]{
            \node[shape=circle,draw,inner sep=0.2pt] (char) {#1};}}

\iclrpreprint
\begin{document}

\maketitle

\begin{abstract}

Developing high-performance CUDA kernels demands specialized knowledge in algorithm implementation, correctness validation, and hardware-aware parallel optimization, creating a substantial expertise barrier and making generating CUDA kernels directly from natural language (Text2CUDA) essential.
Meanwhile, the general-purpose code generation capability of Large Language Models (LLMs) prompts a series of works exploring LLM-based CUDA kernel generation.
They mainly focus on transpilation from high-level frameworks such as PyTorch to CUDA (Torch2CUDA) rather than Text2CUDA, where models must understand the high-level input semantics and handle low-level kernel implementation and validation.
Additionally, these methods are vulnerable to reward hacking due to reliance on predefined test inputs.
In this paper, we propose CUDA-Harness, a framework for harnessing agentic CUDA kernel generation and optimization from natural language. 
Specifically, we introduce Intermediate-Structured Generation to connect high-level semantic understanding with low-level kernel generation. 
To dilute reward hacking in Text2CUDA, we construct Synthesis-Based Verification to provide isolated test data and progressive validation.
Furthermore, we propose Feedback-Adaptive Evolution, a kernel evolution strategy that prioritizes correctness while optimizing performance. 
Finally, through extensive experiments, we demonstrate the effectiveness of CUDA-Harness, with further evaluations illustrating generalization across LLMs, hardware platforms, and to C-to-CUDA transpilation.

\end{abstract}
\section{Introduction}

With the advancement of high-performance computing, CUDA has become the dominant platform for parallel computing, powering applications from artificial intelligence~\citep{pytorch, flash-attention} to quantum computing~\citep{cuda4quantum} and scientific simulation~\citep{cuda4ocean}. 
However, developing high-performance kernels remains challenging, as it requires not only concise knowledge of algorithm implementation and testing details, but also specialized understanding of hardware architectures and parallel optimization. 
These requirements create a substantial expertise barrier that limits the accessibility of GPU programming.
Hence, generating CUDA kernels directly from natural language (Text2CUDA) is essential to bridge the gap between high-level application needs and low-level parallel implementation.

\begin{figure}[t]
    \centering
    \includegraphics[width=\linewidth]{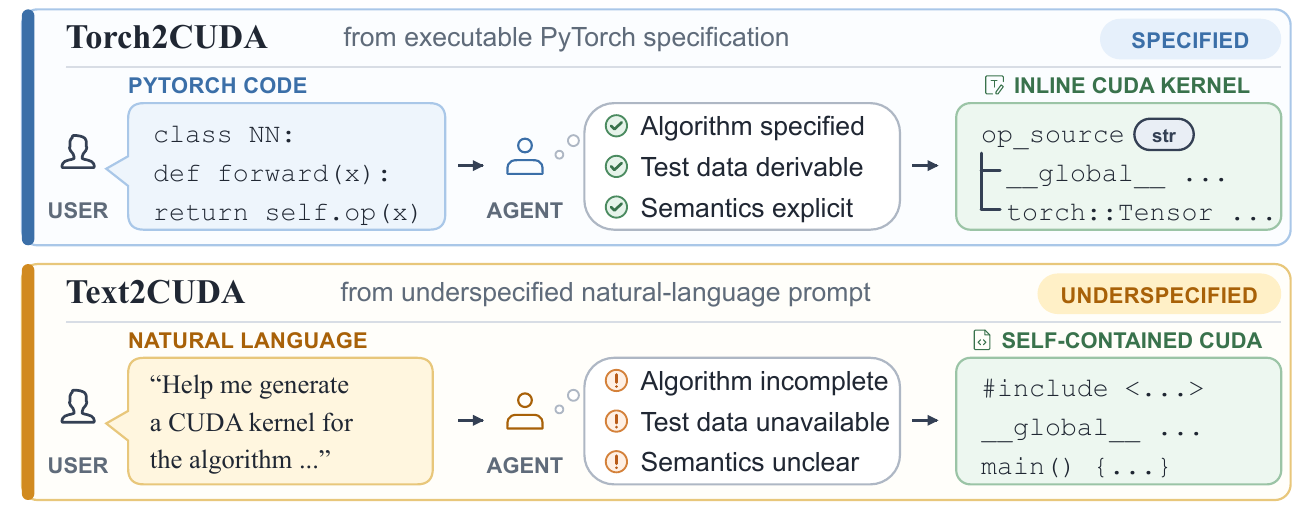}
    \caption{Text2CUDA vs. Torch2CUDA. Torch2CUDA starts from detailed PyTorch code, while Text2CUDA starts from an underspecified natural-language prompt with no available test data.}
    \label{fig:text2cuda}
\end{figure}

Large Language Models (LLMs) have demonstrated remarkable capabilities in general-purpose code generation~\citep{livecodebench-pro}, prompting a series of studies~\citep{llm4cuda-survey} that leverage LLMs to produce high-performance CUDA kernels.
Nevertheless, existing works~\citep{kernelbench, stark} primarily focus on transpilation from high-level frameworks (e.g., Torch2CUDA, which transpiles PyTorch to CUDA) rather than Text2CUDA.
The distinction is fundamental. 
Transpilation starts from programs whose algorithmic details are already specified, whereas Text2CUDA starts from natural language descriptions that can be incomplete and underspecified.
Hence, Text2CUDA requires the model to more deeply understand the high-level intended semantics while simultaneously handling low-level kernel implementation and validation, as illustrated in Fig.~\ref{fig:text2cuda}.
Therefore, a Text2CUDA framework that systematically addresses input semantic understanding is essential.

Existing LLM-based CUDA kernel generation approaches can be broadly categorized into two types: training-based and agent-based. 
Training-based approaches~\citep{cuda-l1, cuda-agent, kernel-smith} enhance the model's CUDA kernel generation capability through supervised fine-tuning or reinforcement learning. 
Nevertheless, they heavily rely on high-quality kernel implementations and optimization trajectories to build the training data~\citep{data-quality}, which is scarce and expensive in practice.
Agent-based approaches~\citep{stark, cudaforge, cupilot} leverage multi-agent collaboration and self-refinement~\citep{self-refine} to achieve training-free kernel generation and optimization.
However, they typically depend on predefined test inputs for validation and feedback construction during iterative refinement, which may lead to reward hacking~\citep{reward-hacking} and optimization over flawed implementations.
Therefore, harnessing agentic CUDA generation is critical for generating correct and efficient CUDA kernels.

In this paper, we propose CUDA-Harness, a framework for harnessing agentic CUDA kernel generation and optimization from natural language.
With three components for kernel generation, verification, and optimization, it connects input semantic understanding with verification-guided refinement to improve correctness and efficiency. 
Empirically, CUDA-Harness shows that natural language can serve as a practical interface for correct and high-performance CUDA kernel development.
The contributions are fourfold.

\begin{itemize}
    \item We introduce Intermediate-Structured Generation, bridging the gap between high-level semantic understanding and low-level kernel synthesis for Text2CUDA.
    \item We construct Synthesis-Based Verification, providing isolated test data synthesis and progressive validation to dilute reward hacking in Text2CUDA.
    \item We propose Feedback-Adaptive Evolution, prioritizing correctness to avoid error accumulation during optimization.
    \item We conduct extensive experiments demonstrating the effectiveness of CUDA-Harness, while extended evaluations illustrate its generalizability across LLMs, hardware platforms, and to C-to-CUDA transpilation.
\end{itemize}
\section{Related Work}

Developing high-performance CUDA kernels remains challenging, as it requires jointly handling algorithmic correctness, verification, hardware-specific constraints, and parallel optimization.
Motivated by these challenges and the strong capabilities of LLMs, LLM-based kernel generation has gained significant attention, which can be broadly divided into training-based and agent-based approaches.

Training-based methods enhance the model's capability to generate high-performance CUDA kernels through large-scale post-training, including supervised fine-tuning (SFT) and reinforcement learning (RL). 
For example, CUDA-L1~\citep{cuda-l1} proposes contrastive RL, using speedup as the reward signal to optimize generation quality.
ConCuR~\citep{concur} generates and curates high-quality kernels with concise reasoning traces to perform SFT. 
QiMeng-Kernel~\citep{qimeng-kernel} employs RL to enable lightweight LLMs providing efficient optimization strategies. 
CUDA-Agent~\citep{cuda-agent} leverages large-scale data synthesis and agentic RL to strengthen the model.
KernelSmith~\citep{kernel-smith} utilizes evolution trajectories as post-training signals to optimize the model as a local improver. 
However, these methods rely on expensive high-quality kernel implementations and optimization traces in practice.

Agent-based methods generate and optimize kernels at test time via multi-agent collaboration and iterative refinement~\citep{self-refine}.
For instance, cuPilot~\citep{cupilot} utilizes roofline-guided prompting for multi-agent kernel optimization. 
STARK~\citep{stark} abstracts expert CUDA engineering into collaborative agents to explore the kernel design space.
ReGraphT~\citep{regrapht} organizes historical CUDA optimization trajectories as a graph to perform efficient searches.
CudaForge~\citep{cudaforge} adopts a two-agent loop with profiling tools to iteratively refine generated kernels. 
Nevertheless, due to dependence on predefined test inputs to guide optimization, they are typically vulnerable to reward hacking.

Although existing works have achieved notable success, they primarily focus on transpilation such as Torch2CUDA rather than the more general Text2CUDA. 
For Text2CUDA, CUDA-LLM~\citep{cuda-llm} provides an agent-based framework for iteratively improving kernel implementations under validation, while CUDABench~\citep{cudabench} evaluates the Text2CUDA capabilities of LLMs across diverse domains. 
However, these efforts remain constrained by the limitations of existing LLM-based kernel generation methods. 
Therefore, a framework for harnessing agentic Text2CUDA is essential.

\section{Overview}

As the self-contained CUDA program serves as the basic unit providing the complete context to compile, execute, and validate a kernel, we analyze its composition and execution model and present our CUDA-Harness in this section.

\subsection{Anatomy of Self-Contained CUDA}
\label{sec:anatomy-cuda}

\begin{wrapfigure}{r}{0.67\textwidth}
    \centering
    \vspace{-20pt}
\begin{cudacode}
#include <cuda_runtime.h>

__global__ void vecAddInplace(float* a, float* b, int n) {
    int idx = blockIdx.x * blockDim.x + threadIdx.x;
    if (idx < n) a[idx] += b[idx];
}

int main() {
    int n = 1024; // define number of elements
    ... // prepare dummy inputs or load test inputs
    ... // allocate device memory and copy inputs to device

    // prepare launch config and launch the kernel
    int blocksPerGrid = (n + 255) / 256; // launch config
    vecAddInplace<<<blocksPerGrid, 256>>>(a, b, n);
    
    ... // copy result to host and post-process if needed
    ... // verify results if needed and free memory
    return 0;
}
\end{cudacode}
    \vspace{-12pt}
    \caption{A simplified self-contained CUDA program.}
    \label{fig:self-contained-cuda}
    \vspace{-12pt}
\end{wrapfigure}

A self-contained CUDA program typically comprises necessary headers, device-side kernels and host-side code.
Fig.~\ref{fig:self-contained-cuda} showcases a simplified example of the self-contained CUDA program.
Specifically, the device-side kernels perform memory accesses and execute the parallel computation, while the host-side code prepares the inputs, manages the memory, and launches kernels with appropriate launch configurations.

Beyond the structure, the execution depends on well-prepared test inputs for validation and further performance measurement.
Additionally, as problem scale and input dimensions directly influence the performance bottlenecks observed in CUDA kernels at runtime, dummy inputs are insufficient.

In Torch2CUDA, the input PyTorch code already reveals algorithmic detail and naturally provides a basis for test construction.
However, Text2CUDA takes only high-level but often underspecified natural language intents as input, lacking the test data needed for validation.

\begin{figure*}[t]
    \centering
    \includegraphics[width=\linewidth]{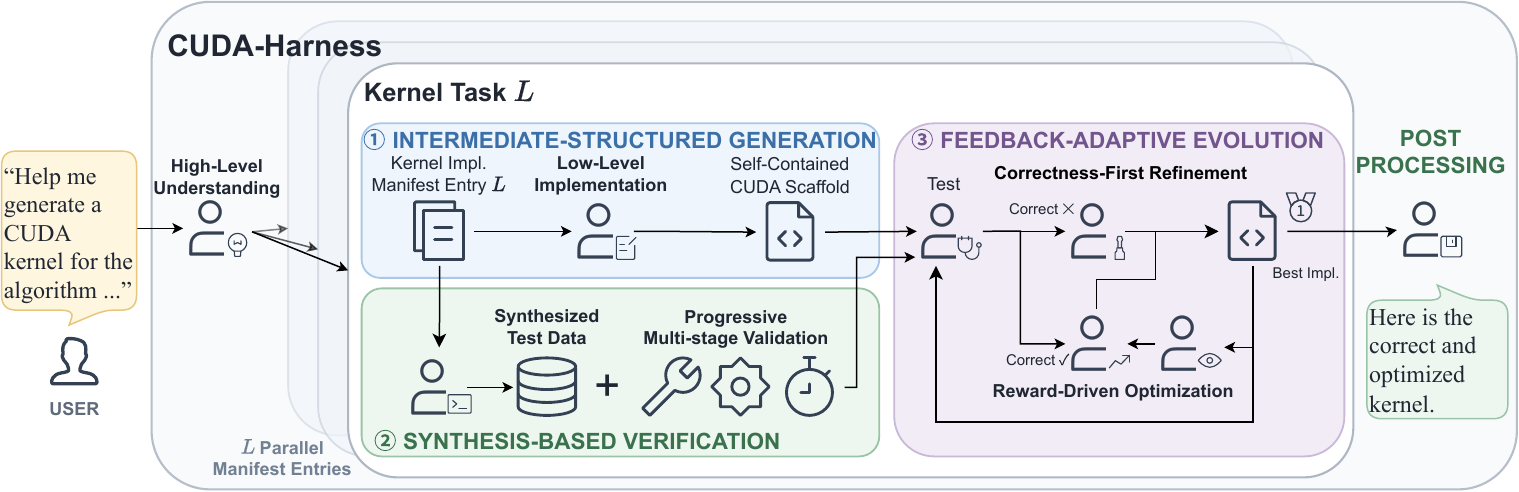}
    \caption{The overview of the CUDA-Harness.}
    \label{fig:cuda-harness}
\end{figure*}

\subsection{From Anatomy to CUDA-Harness}

According to the anatomy of the self-contained CUDA program, the challenges of Text2CUDA are:

\noindent \textit{
\textbf{Challenge 1 {[Bridging Intent and Implementation]}}: 
How to enable the agent to deeply understand the high-level natural language intents and complete the missing details is the first challenge.
In addition, how to use the completed information to guide the agent towards easily testable kernel implementations is another challenge.
}

\noindent \textit{
\textbf{Challenge 2 {[Missing Test Data for Validation]}}:
How to synthesize test data for kernel validation is the first challenge.
Furthermore, how to avoid reward hacking of kernel implementation due to test data synthesis is another challenge.
}

We propose the CUDA-Harness to address the above challenges.
The overview of the CUDA-Harness is presented in Fig.~\ref{fig:cuda-harness}.
\circled{1} Intermediate-Structured Generation introduces two intermediate blueprints, namely the kernel implementation manifest and the self-contained CUDA scaffold.
Given a natural language input, the agent first produces the implementation manifest to externalize semantic understanding and implementation planning.
For each entry in the manifest, the agent then instantiates the CUDA scaffold and focuses on the concrete implementation of the corresponding kernel.
\circled{2} Synthesis-Based Verification decouples test data synthesis from kernel generation, allowing the agent to construct test inputs in an isolated environment for validation.
Based on the feedback from the progressive validation, \circled{3} Feedback-Adaptive Evolution optimizes the kernel for better performance while preserving correctness.
Once all manifest entries are implemented, CUDA-Harness collects the best-performing kernels together with the corresponding kernel launchers for downstream post-processing.

\section{CUDA-Harness}
\label{sec:cuda-harness}

This section presents the components of CUDA-Harness.

\subsection{Intermediate-Structured Generation}
\label{sec:generation}

Within Intermediate-Structured Generation, the kernel implementation manifest enhances high-level semantic understanding, while the self-contained CUDA scaffold supports low-level kernel implementation.

\subsubsection{Kernel Implementation Manifest}

The kernel implementation manifest is a structured per-kernel specification, where each entry stands alone.
Each entry records the essential information to implement a kernel, including the kernel name, the input and output descriptions of the kernel, a functional description that reveals algorithmic detail, and a language-agnostic reference code snippet.
Additionally, each entry carries detailed test information, encompassing the shapes and data types of the test inputs and outputs.

When the agent produces the manifest from the natural language input, it does not focus on writing code for concrete kernel implementation. 
Therefore, the agent can concentrate on understanding the high-level semantics and on completing details that the original input leaves underspecified.

\subsubsection{Self-Contained CUDA Scaffold}
\label{sec:scaffold}

\begin{wrapfigure}{r}{0.69\textwidth}
    \centering
    \vspace{-20pt}
\begin{cudacodearg}[basicstyle=\ttfamily\footnotesize]
[headers]
[kernel and launcher declaration]
[launcher]
[kernel]
T* read_binary_file(...) { ... }
T* write_binary_file(...) { ... }
int main(int argc, char **argv) {
    #pragma message("Calling convention: ...")
    [check arguments and read test inputs]
    [invoke the launcher function]
    [write kernel outputs]
}
\end{cudacodearg}
    \vspace{-12pt}
    \caption{A simplified self-contained CUDA scaffold.}
    \label{fig:cuda-scaffold}
    \vspace{-12pt}
\end{wrapfigure}

The self-contained CUDA scaffold is a code template that tells the agent how to implement the self-contained CUDA program based on the manifest entry.
Fig.~\ref{fig:cuda-scaffold} illustrates a simplified example of the self-contained CUDA scaffold.
The scaffold includes placeholders for the components of the self-contained CUDA program illustrated in Sec.~\ref{sec:anatomy-cuda}.

Within the scaffold, we aggregate the operations related to kernel launch, including memory management, data transfer, and launch configuration definition, into a dedicated launcher function rather than scattering it across the host-side code.
With this arrangement, the agent can concentrate on the kernel and its launcher and ignore the remaining host-side details.
However, because the launch configuration determines the memory-access pattern of the kernel, the order in which kernel and launcher are generated matters.
During initial kernel generation, the agent assumes a reasonable memory access pattern.
However, when generating the launcher, the launch configuration is specified. 
When the implicitly assumed memory access pattern in the kernel conflicts with the explicitly defined launch configuration in the launcher, the agent will reflect~\citep{reflection} and retrospectively edit the already generated kernel, producing invalid CUDA programs.
Therefore, 

Moreover, as an initial step towards mitigating reward hacking, the scaffold prohibits the use of dummy inputs and provides helper functions for reading the well-prepared test inputs.
The test input paths are passed as command-line arguments, and the calling convention is emitted at compile time.
Thus, the agent remains unaware of the details of the test inputs when implementing the kernel and cannot tailor an overfitted implementation to them.

\subsection{Synthesis-Based Verification}
\label{sec:verification}

To address the lack of test data in Text2CUDA, Synthesis-Based Verification enables decoupled test data synthesis and evaluates the correctness and performance through progressive multi-stage validation.
Rather than theoretically eliminating the risk of reward hacking, the purpose of Synthesis-Based Verification is to provide a trustworthy correctness signal for kernel refinement when benchmark test data is unavailable.

\subsubsection{Decoupled Test Data Synthesis}

Although the self-contained CUDA scaffold eliminates a pathway for reward hacking by using well-prepared test inputs, there remain risks of reward hacking during test data synthesis.
If test data synthesis shares context with kernel generation, the agent may adapt test inputs and reference outputs to the kernel implementation rather than to the intended algorithm.
Therefore, we decouple test data synthesis and kernel generation.
Specifically, the test inputs and reference outputs are produced through standard numerical libraries (e.g., NumPy and PyTorch) in an isolated environment, where context isolation is enforced. Namely, kernel generation and test data synthesis use separate contexts to avoid cross-contamination.

However, decoupling test data synthesis is insufficient, as reward hacking can still arise through the validation rule due to how the test data is synthesized.
The functional validation for a kernel $\mathcal{K}$ under test, with test inputs $x$ and reference outputs $y$, is defined as
\begin{equation}
\label{eq:validation}
\mathcal{V}(\mathcal{K}, x, y) = \mathbb{I}
\Big[
\lvert \mathcal{K}(x) - y \rvert 
\leqslant \epsilon_{\text{atol}} + \epsilon_{\text{rtol}}\lvert y \rvert
\Big]
\end{equation}
where $\mathcal{K}(x)$ denotes the kernel outputs, $\mathbb{I}$ is the indicator function, $\epsilon_{\text{atol}}$ and $\epsilon_{\text{rtol}}$ are the absolute and relative tolerances, respectively.
According to Eq.~\eqref{eq:validation}, when the test inputs concentrate on small-magnitude values, the output magnitude $\lvert y \rvert$ also tends to be small, causing $\epsilon_{\text{rtol}}\lvert y \rvert$ to diminish and $\epsilon_{\text{atol}}$ to dominate the criterion.
In such cases, a numerically incorrect kernel may still pass the validation, leading to reward hacking when this feedback is used for optimization.

Hence, we require the agent to draw test inputs from a broad numeric range.
For instance, test inputs following a Gaussian distribution with a mean of 5.0 and a standard deviation of 10.0 are required to induce distribution shift and cover broader numeric ranges than dummy or small-magnitude inputs.
However, larger-magnitude inputs may amplify floating-point accumulation error, which makes validation stricter and can increase false negatives (i.e., rejecting correct kernels).
Nevertheless, we adopt this trade-off deliberately to reduce false positives (i.e., accepting incorrect kernels), since accepting an incorrect kernel would mislead subsequent optimization.

\subsubsection{Progressive Multi-Stage Validation}
\label{sec:progressive-validation}

Given synthesized test data, CUDA-Harness employs a progressive multi-stage protocol for kernel validation.
As validation stages are naturally ordered, with performance mattering only after a kernel compiles and produces correct outputs, CUDA-Harness rejects invalid kernels early to avoid unnecessary runs on target hardware.
Specifically, CUDA-Harness follows a fail-fast order that checks compilation first, then functional correctness, and finally runtime performance.

To support the validation protocol, CUDA-Harness exposes a validation toolkit based on the Model Context Protocol (MCP)~\citep{mcp}.
These tools provide a uniform interface to the agent, while their concrete execution remains bound to the environment where the toolkit is deployed, enabling hardware-scalable validation.
The toolkit comprises tools that handle compilation, functional validation, and performance profiling.

The compilation tool checks whether a kernel $\mathcal{K}$ can be built successfully under a given compilation setting, which can be represented as
\begin{equation}
\label{eq:compilation}
\mathcal{C}(\mathcal{K}; \phi) = \mathbb{I}
\Big[
\mathcal{K}\text{ compiles under }\phi
\Big]
\end{equation}
where $\phi$ is the compilation parameters, such as \texttt{"-arch=sm\_80"}.
Given the test inputs $x$ and reference outputs $y$, the functional validation tool verifies whether the kernel outputs match the reference, producing the result $\mathcal{V}(\mathcal{K}, x, y)$.
Utilizing profilers such as Nsight Systems and Nsight Compute, the performance profiling tool can measure the runtime performance $\mathcal{P}(\mathcal{K}, x)$, where a larger $\mathcal{P}(\cdot)$ indicates a better performance.
Taking latency as an example, the performance can be defined as
\begin{equation}
\label{eq:profiling}
\mathcal{P}_{\text{latency}}(\mathcal{K}, x) = 1  \text{ / Latency}(\mathcal{K}, x)
\end{equation}

Integrating these tools, the agent first invokes the compilation tool and is informed of the calling convention emitted by the scaffold via a compile-time \texttt{\#pragma message} as described in Sec.~\ref{sec:scaffold}.
When the kernel passes compilation, the agent calls the functional validation tool based on the calling convention, followed by the performance profiling tool upon successful validation.

\subsection{Feedback-Adaptive Evolution}
\label{sec:evolution}

To optimize the kernel while preserving correctness, Feedback-Adaptive Evolution combines correctness-first refinement and reward-driven optimization to perform test-time kernel evolution.

\subsubsection{Correctness-First Refinement}
\label{sec:correct-refinement}

As performance only becomes meaningful after correctness is in place, the agent must care about whether the generated kernel compiles and produces the correct outputs.
However, the agent tends to introduce advanced and aggressive optimization strategies before the implementation is fully stable and the correctness is guaranteed.
When optimization is built on a kernel that looks high-performance but is functionally flawed, subsequent refinement typically remains trapped in the same faulty pattern. 
Therefore, CUDA-Harness introduces correctness-first refinement into test-time kernel evolution.

The refinement branches based on which validation stage reports the failure.
If the generated kernel fails to compile, the agent is steered towards a specific repair based on the error details provided by the compilation tool.
When the kernel fails functional validation, the agent is instructed to retreat to a more conservative implementation, prioritizing correctness above all else. 

Through correctness-first refinement, once the kernel is reset to a simpler implementation that compiles and passes functional validation, correctness is initially guaranteed.
Therefore, subsequent performance optimization can proceed from a trustworthy and well-behaved base.

\subsubsection{Reward-Driven Optimization}
\label{sec:performance-optimizatin}

Building on correctness-first refinement, CUDA-Harness performs reward-driven optimization for the test-time kernel evolution.
Benefiting from the progressive multi-stage validation in Sec.~\ref{sec:progressive-validation}, there is a natural and verifiable reward signal characterizing kernel quality.
Since the signal comes from compilation, functional validation, and profiling on the target hardware, it is measurable and trustworthy rather than estimated.
Therefore, we employ Reinforcement Learning with Verifiable Reward (RLVR)~\citep{rlvr} to drive optimization.
With compilation parameters $\phi$, test inputs $x$, and reference outputs $y$ fixed, the results of compilation, functional validation, and performance profiling for a generated kernel $\mathcal{K}$ can be simplified as $\mathcal{C}(\mathcal{K})$, $\mathcal{V}(\mathcal{K})$ and $\mathcal{P}(\mathcal{K})$, respectively.
Hence, the reward signal can be defined as
\begin{equation}
\label{eq:rlvr-reward}
r(\mathcal{\mathcal{K}}) = -1 + \mathcal{C}(\mathcal{K}) + \mathcal{C}(\mathcal{K})\mathcal{V}(\mathcal{K})\mathcal{P}(\mathcal{K})
\end{equation}
Concretely, a penalty is applied when the generated kernel fails to compile.
If the kernel compiles and passes functional validation, the performance determines the reward, with higher performance yielding a higher reward.

However, there are no updatable parameters to drive the optimization.
Hence, we introduce optimization insight summarization, in which the agent extracts and updates the experience from what changed and how the reward moved to steer the optimization.
Given the pre-optimization kernel $\mathcal{K}_{\text{pre}}$ and the optimized kernel $\mathcal{K}_{\text{post}}$, the optimization insight summarization process is defined as
\begin{equation}
\label{eq:insight-summarization}
e = \mathcal{M}_e(\text{Diff}(\mathcal{K}_{\text{post}}, \mathcal{K}_{\text{pre}}), r(\mathcal{K}_{\text{post}}) - r(\mathcal{K}_{\text{pre}}))
\end{equation}
where the agent $\mathcal{M}_e$ analyzes the difference in implementation and reward, and distills the optimization experience $e$, such as which optimization strategies are effective or ineffective.

Thus, given $N$ rounds of iterative optimization and $G$ generated kernels per round, and assuming the accumulated optimization insights at the $i$-th round are $E_{1:i}$, reward-driven optimization is written as the following objective:
\begin{equation}
\label{eq:rlvr-object}
\mathcal{J}(E_i) = \max \mathbb{E}_{\{\mathcal{K}_{i+1, j}\}_{j=1}^G \sim \mathcal{M}(\cdot \mid \mathcal{K}_i, E_{1: i})} \left[ \Delta r_{i, j} \right]
\end{equation}
where $\mathcal{M}$ is the optimization agent, $\mathcal{K}_i$ is the best kernel at the $i$-th round, $\mathcal{K}_{i+1, j}$ is the $j$-th generated kernel for round $i+1$, $\Delta r_{i, j} = r(\mathcal{K}_{i+1, j}) - r(\mathcal{K}_i)$ is the reward difference, and $E_{1:i} = \{E_1, \ldots, E_i\}$ accumulates the optimization insights from round 1 to $i$.
The set $E_i$ collects the $G$ optimization insights from round $i$, namely $E_i = \{ \mathcal{M}_e(\text{Diff}(\mathcal{K}_{i-1, j}, \mathcal{K}_{i, j}), r(\mathcal{K}_{i, j}) - r(\mathcal{K}_{i-1, j})) \}_{j=1}^G$.
\section{Evaluation}

In this section, the experimental setup is first introduced.
The comparison results and ablation studies are presented in Sec.~\ref{sec:eval-main-results}.
The generalizability of CUDA-Harness is evaluated in Sec.~\ref{sec:eval-generalization} under cross-LLM, cross-hardware, and C-to-CUDA transpilation scenarios.

\begin{table*}[t]
\centering
\caption{The comparison and ablation results. CUDA-ISG retains only Intermediate-Structured Generation, while CUDA-Harness* further removes Correctness-first Refinement from CUDA-Harness.}
\label{tab:comparison}
\resizebox{\textwidth}{!}{
\begin{tabular}{@{}ccccccccccccc@{}}
\toprule
\multirow{2}{*}{Method} & \multicolumn{3}{c}{Overall}                    & \multicolumn{3}{c}{Level 1}                    & \multicolumn{3}{c}{Level 2}                     & \multicolumn{3}{c}{Level 3}                    \\ \cmidrule(l){2-13} 
                        & Comp.         & Func.         & RScore         & Comp.         & Func.         & RScore         & Comp.          & Func.         & RScore         & Comp.         & Func.         & RScore         \\ \midrule
Baseline                & 90.1          & 60.9          & 80.1           & 89.6          & 69.0          & 87.7           & 92.6           & 66.2          & 86.1           & 88.2          & 47.6          & 66.6           \\
OpenCode                & 92.1          & 60.6          & 78.7           & 93.6          & 66.6          & 82.8           & 92.0           & 65.8          & 81.9           & 90.8          & 49.4          & 71.3           \\
Codex                   & 98.3          & 69.5          & 86.5           & 99.2          & 79.2          & 91.3           & 98.6           & 75.4          & 94.5           & 97.2          & 54.0          & 73.6           \\
KernelSkill             & 99.4          & 63.7          & 86.1           & 99.0          & 72.0          & 94.4           & 100.0          & 70.4          & 94.3           & 99.2          & 48.6          & 69.7           \\
CudaForge               & 99.8          & 67.7          & 87.9           & 99.6          & 75.2          & 92.7           & 99.8           & 73.2          & 93.7           & 100.0         & 54.8          & 77.4           \\
CUDA-ISG                & 94.5          & 63.9          & 84.4           & 93.4          & 72.0          & 91.5           & 96.4           & 69.2          & 91.1           & 93.6          & 50.4          & 70.6           \\
CUDA-Harness*           & 99.3          & 70.8          & 94.9           & 99.6          & 79.0          & 100.6          & 99.6           & 75.8          & 101.3          & 98.6          & 57.6          & 82.8           \\
\textbf{CUDA-Harness}   & \textbf{99.7} & \textbf{80.5} & \textbf{110.7} & \textbf{99.4} & \textbf{86.2} & \textbf{116.2} & \textbf{100.0} & \textbf{84.8} & \textbf{115.6} & \textbf{99.8} & \textbf{70.4} & \textbf{100.4} \\ \bottomrule
\end{tabular}
}
\end{table*}

\subsection{Experimental Setup}

We evaluate on CUDABench~\citep{cudabench}, which comprises three difficulty levels of prompts that progressively remove details.
As CUDA-Harness is decoupled from the underlying LLM, we use Seed2.0 Lite (\texttt{doubao-seed-2-0-lite-260215})~\citep{seed2} with thinking mode disabled as the invoked LLM.
The model is fixed throughout the evaluation, isolating the effect of the harness from variation across LLMs.
For each kernel, generation is performed only once, while repetitions are used solely during the performance-measurement stage.
The experiments are conducted on an NVIDIA A40 GPU, which features the Ampere architecture and \texttt{sm\_80} compute capability.

\subsubsection{CUDA-Harness Setup}

The agent operates under the reasoning-action (ReAct)  paradigm~\citep{react}.
The number of iterative optimization rounds is set to $N = 3$, with $G = 3$ kernels generated per round.
Leveraging Nsight Systems, the performance profiling tool measures the execution time of the kernel over multiple runs and uses the average latency to compute the performance metric $\mathcal{P}(\cdot) = \mathcal{P}_{\text{latency}}(\cdot)$, as defined in Eq.~\eqref{eq:profiling}.
Following the evolution process detailed in Sec.~\ref{sec:evolution}, optimization thus proceeds towards a correct and fast kernel.

\subsubsection{Evaluation Metrics}

Using the compilation parameters and test data provided by the benchmark, we report two correctness metrics, the compilation success rate (Comp.) and functional correctness rate (Func.).
Since CUDA-Harness uses execution time to guide test-time evolution, for a fair comparison, we define the runtime performance score (RScore) as $1/\texttt{execution time (ms)}$ computed over multiple measurements, which is not a percentage. The average RScore is reported, with kernels failing the correctness validation assigned a RScore of 0.

\subsection{Main Results}
\label{sec:eval-main-results}

\subsubsection{Comparison with Baselines}

We compare CUDA-Harness with the official CUDABench~\citep{cudabench} evaluation procedure, which serves as the baseline harness.
Additionally, we further compare the general harnesses,  OpenCode~\citep{opencode} and Codex~\citep{codex}, and agentic Torch2CUDA systems, KernelSkill~\citep{kernelskill} and CudaForge~\citep{cudaforge}.
The results are presented in Tab.~\ref{tab:comparison}.
Across all difficulty levels, CUDA-Harness outperforms the baseline in terms of compilation success rate, functional correctness rate, and the performance score, demonstrating the effectiveness of CUDA-Harness.

As the difficulty level increases, the performance gap between the baseline and CUDA-Harness grows, where higher levels indicate less information in the prompt.
Since the baseline generates kernels directly from the raw prompt, it is forced to infer the high-level intents while simultaneously handling low-level implementation within a single generation pass.
In contrast, CUDA-Harness introduces Intermediate-Structured Generation to complete underspecified details, allowing it to remain competitive even at Level 3.

\subsubsection{Ablation Study}

To isolate the contribution of each component, we incrementally enable them in Tab.~\ref{tab:comparison}.

CUDA-ISG retains only Intermediate-Structured Generation (Sec.~\ref{sec:generation}).
CUDA-ISG outperforms the baseline, demonstrating the effectiveness and the necessity of the proposed intermediate blueprints.

CUDA-Harness* adds Synthesis-Based Verification (Sec.~\ref{sec:verification}) and Reward-Driven Optimization (Sec.~\ref{sec:performance-optimizatin}) on top of CUDA-ISG while withholding Correctness-First Refinement (Sec.~\ref{sec:correct-refinement}).
Compared with CUDA-ISG, its higher compilation success rate and performance score shows that Reward-Driven Optimization is effective at repairing compilation failures and optimizing the generated kernels.
However, the remaining gap to the full CUDA-Harness reveals why Correctness-First Refinement matters.
Through Correctness-First Refinement, CUDA-Harness achieves a higher functional correctness rate than CUDA-Harness*, further confirming the importance of correctness-first refinement in preserving correctness throughout the optimization process.

\subsubsection{Efficacy of the Synthesis-Based Verification}

We evaluate whether the proposed Synthesis-Based Verification can provide a trustworthy correctness signal for CUDA-Harness.
Since the benchmark test data is unavailable during test-time evolution, the synthesized data should lead to validation results that are consistent with the benchmark validation results.
To examine the consistency, Tab.~\ref{tab:test-data-synthesis} compares the validation results obtained using synthesized test data with those obtained using benchmark test data.

\begin{wraptable}{r}{0.455\textwidth}
    \centering
    \vspace{-12pt}
    \caption{The consistency between Synthesis-Based Verification and benchmark validation.}
    \label{tab:test-data-synthesis}
    \vspace{-8pt}
\resizebox{\linewidth}{!}{
\begin{tabular}{@{}ccc@{}}
\toprule
\multirow{2}{*}{\begin{tabular}[c]{@{}c@{}}Synthesis-based\\ Verification\end{tabular}} & \multicolumn{2}{c}{Benchmark Validation} \\ \cmidrule(l){2-3} 
                                                                                        & Pass                & Fail               \\ \midrule
Pass                                                                                    & \textbf{73.6\%}     & 8.7\%              \\
Fail                                                                                    & 6.9\%               & \textbf{10.8\%}    \\ \bottomrule
\end{tabular}
}
    \vspace{-12pt}
\end{wraptable}

As shown in Tab.~\ref{tab:test-data-synthesis}, Synthesis-Based Verification is largely consistent with benchmark validation.
Treating benchmark validation as the reference and Synthesis-Based Verification as the prediction, Table~\ref{tab:test-data-synthesis} shows an accuracy of 84.4\%, with precision 89.4\% and recall 91.4\%.
These results indicate that Synthesis-Based Verification provides a strong correctness signal for test-time evolution.

Furthermore, we manually analyze the false positives (8.7\% of all cases) where Synthesis-Based Verification passes while benchmark validation fails.
Among these false positive cases, 38.1\% stem from prompt misinterpretation (e.g., interpreting $A^3$ as element-wise cubing), which misleads the subsequent test data synthesis and kernel generation.
Another 30.5\% arise from omitting steps during benchmark-specific post-processing (e.g., averaging the kernel outputs after execution).
The remaining 31.4\% are caused by a defect in the CUDABench template that loads test data as \texttt{float}, causing incorrect loading for underlying data types such as \texttt{uint8} and \texttt{int32}.

\subsection{Generalization Analysis}
\label{sec:eval-generalization}

\begin{table*}[t]
\centering
\caption{The generalization results of CUDA-Harness across LLMs and across hardware platforms on CUDABench.}
\label{tab:generalization-llm-hardware}
\begin{tabular}{@{}cccccccc@{}}
\toprule
\multirow{2}{*}{Method}                & \multirow{2}{*}{Metric} & \multicolumn{3}{c}{Cross-LLMs}                                                                                                                                                             & \multicolumn{3}{c}{Cross-Hardware}                                                                                                \\ \cmidrule(l){3-8} 
                                       &                         & \begin{tabular}[c]{@{}c@{}}Seed\\ 2.0 Lite\end{tabular} & \begin{tabular}[c]{@{}c@{}}DeepSeek\\ V3.2\end{tabular} & \multicolumn{1}{c|}{\begin{tabular}[c]{@{}c@{}}GLM\\ 5.1\end{tabular}} & A40            & \begin{tabular}[c]{@{}c@{}}1660\\ SUPER\end{tabular} & \begin{tabular}[c]{@{}c@{}}Jetson\\ AGX Orin\end{tabular} \\ \midrule
\multirow{3}{*}{Baseline}              & Comp.                   & 90.1                                                    & 96.5                                                    & 94.8                                                                   & 90.1           & 90.5                                                 & 90.2                                                      \\
                                       & Func.                   & 60.9                                                    & 57.8                                                    & 60.7                                                                   & 60.9           & 60.3                                                 & 59.9                                                      \\
                                       & RScore                  & 80.1                                                    & 79.0                                                    & 79.7                                                                   & 80.1           & 52.7                                                 & 29.9                                                      \\ \midrule
\multirow{3}{*}{\textbf{CUDA-Harness}} & Comp.                   & \textbf{99.7}                                           & \textbf{99.6}                                           & \textbf{99.2}                                                          & \textbf{99.7}  & \textbf{99.7}                                        & \textbf{99.7}                                             \\
                                       & Func.                   & \textbf{80.5}                                           & \textbf{76.3}                                           & \textbf{101.8}                                                         & \textbf{80.5}  & \textbf{79.8}                                        & \textbf{79.4}                                             \\
                                       & RScore                  & \textbf{110.7}                                          & \textbf{80.1}                                           & \textbf{107.8}                                                         & \textbf{110.7} & \textbf{73.9}                                        & \textbf{42.4}                                             \\ \bottomrule
\end{tabular}
\end{table*}

\subsubsection{Generalization across LLMs}

As CUDA-Harness is decoupled from the underlying LLM, we further evaluate whether it generalizes across different LLMs.
Besides Seed2.0 Lite~\citep{seed2}, we invoke DeepSeek-V3.2~\citep{deepseekv32} and GLM-5.1~\citep{glm5} as the underlying LLMs, and compare CUDA-Harness with the baseline harness on CUDABench using an NVIDIA A40 GPU.
As shown in Tab.~\ref{tab:generalization-llm-hardware}, CUDA-Harness consistently outperforms the corresponding baseline across all three LLMs in terms of compilation success rate, functional correctness rate, and runtime performance score.
These results substantiate the LLM-agnostic design of CUDA-Harness, as its improvements do not rely on a specific underlying LLM.

\subsubsection{Generalization across Hardware}

We evaluate CUDA-Harness using Seed2.0 Lite on three GPUs with distinct architectures and compute capabilities, including the NVIDIA A40, the NVIDIA GeForce GTX 1660 SUPER, and the NVIDIA Jetson AGX Orin.
The benchmark and the utilized LLM remain identical to Sec.~\ref{sec:eval-main-results}.
As shown in Tab.~\ref{tab:generalization-llm-hardware}, the experimental results vary across hardware platforms due to differences in software (e.g., compilers) versions and hardware compute capabilities.
However, CUDA-Harness maintains close compilation success rates and functional correctness rates across the three platforms.
Although the performance score changes with the hardware compute capability, CUDA-Harness consistently outperforms the baseline on each platform.
These results demonstrate that CUDA-Harness is not tailored to a single hardware environment and can generalize across different hardware platforms.

\subsubsection{Generalization to C-to-CUDA}

We further evaluate under the C-to-CUDA transpilation scenario, where the input is sequential C code rather than natural language, and we evaluate on BabelTower~\citep{babeltower, qimeng-mupa}.
C code provides comprehensive semantics, including test inputs and outputs descriptions and algorithmic implementation details.
For CUDA-Harness, the number of iterative optimization rounds is set to $N = 1$, with $G = 1$ kernels generated per round.

\begin{wraptable}{r}{0.45\textwidth}
    \centering
    \vspace{-12pt}
    \caption{The generalization results of CUDA-Harness to C-to-CUDA transpilation.}
    \label{tab:generalization-c2cuda}
    \vspace{-8pt}
\resizebox{\linewidth}{!}{
\begin{tabular}{@{}cccc@{}}
\toprule
Method                & Comp.         & Func.         & Score          \\ \midrule
Baseline              & 94.8          & 90.1          & 451.6          \\
\textbf{CUDA-Harness} & \textbf{98.7} & \textbf{97.1} & \textbf{514.8} \\ \bottomrule
\end{tabular}
}
    \vspace{-12pt}
\end{wraptable}

As shown in Tab.~\ref{tab:generalization-c2cuda}, CUDA-Harness outperforms the baseline, even though the evolution budget is relatively small, illustrating that CUDA-Harness can easily generalize to C-to-CUDA transpilation well.
Since Intermediate-Structured Generation is not tied to natural language input, decoupling understanding and implementation helps the agent capture the C code and generate the CUDA kernel.

\section{Conclusion}

In this paper, we propose CUDA-Harness, a framework for harnessing agentic CUDA kernel generation and optimization from natural language.
To address the challenges in Text2CUDA, we introduce Intermediate-Structured Generation to bridge the gap between high-level intents and low-level implementation.
We develop Synthesis-Based Verification, providing isolated test data synthesis and progressive validation.
We propose Feedback-Adaptive Evolution to prioritize correctness while optimizing kernel performance during test-time evolution.
Through extensive experiments, we demonstrate the effectiveness of CUDA-Harness.
Additionally, we illustrate that CUDA-Harness generalizes across diverse hardware platforms and extends effectively to C-to-CUDA transpilation.

\bibliography{main}
\bibliographystyle{iclr_arxiv}

\end{document}